\documentclass[11pt,a4paper]{article}
\usepackage[hyperref]{acl2020}
\usepackage{times}
\usepackage{latexsym}

\usepackage{microtype}

\aclfinalcopy 
\def\aclpaperid{7} 

\usepackage{booktabs}
\usepackage{graphicx}
\usepackage{comment}
\usepackage{multirow}
\usepackage{stackrel}
\usepackage{amssymb}
\usepackage{amsmath}
\usepackage{float}
\usepackage{color}
\usepackage{xcolor}

\title{Stance Prediction for Contemporary Issues: Data and Experiments}

\author{ Marjan Hosseinia \\
University of Houston\\
\texttt{mhosseinia@uh.edu} \\ \And
Eduard Dragut\\
Temple University\\
\texttt{edragut@temple.edu} \And
Arjun Mukherjee \\
University of Houston \\
\texttt{arjun@cs.uh.edu}\\
}

\date{}

\begin{document}
\maketitle
\begin{abstract}
We investigate whether pre-trained bidirectional transformers with sentiment and emotion information improve stance detection in long discussions of contemporary issues.  As a part of this work, we create a novel stance detection dataset covering $419$ different controversial issues and their related pros and cons collected by procon.org in nonpartisan format. Experimental results show that a shallow recurrent neural network with sentiment or emotion information can reach competitive results compared to fine-tuned BERT with $20\times$ fewer parameters. We also use a simple approach that explains which input phrases contribute to stance detection. 
\end{abstract}

\section{Introduction}

Stance detection identifies whether an opinion is in favor of an idea or opposes it. It has a tight connection with sentiment analysis; however, stance detection usually investigates the two-sided relationship between an opinion and a question. For example, `should abortion be legal?' or `is human activity primarily responsible for global climate change?'


Contemporary debatable issues, even though non-political, usually carry some political weight and controversy. For example, legislators may allow soda vending machines in our school or consider obesity as a health issue that directly impacts soda manufacturers and insurance companies respectively. On a larger scale, an issue such as climate change is being discussed in US presidential debates constantly. Meanwhile, information about these issues is mostly one-sided and provided by left or right partisan resources. Such information forms public beliefs, has persuasive power, and promotes confirmation bias \cite{StanojevicADO19}, the humans' tendency to search for the information which confirms their existing beliefs \footnote{\url{www.procon.org/education.php}}.  Confirmation bias permits internet debates and promote discrimination, misinformation, and hate speech, all of which are emerging problems in user posts of social media platforms.  
\par
Although there are many attempts to automatic identification and removal of such contents from online platforms, the need for accessing bi-partisan information that cultivates critical thinking and avoids confirmation bias remains. In this regard, a few web sources, such as procon.org, present information in a non-partisan format and being used as a resource for improving critical thinking in educational training by teachers \footnote{\url{https://www.procon.org/view.background-resource.php?resourceID=004241}}. 

Here, we aim to improve such resources by automatic stance detection of pro or con-perspectives regarding a debatable issue. We extend our previous work \cite{hosseinia2019pro} by  creating a new dataset from procon.org  with  $419$ distinct issues and their two-sided perspectives annotated by its experts \footnote{\url{https://github.com/marjanhs/procon20/}}. Then, we leverage external knowledge to identify the stance of a perspective towards an issue that is mainly represented in the form of a question. 

The latest progress in pre-trained language models \cite{howard-ruder-2018-universal} and transformers \cite{devlin-etal-2019-bert,yang2019xlnet}  
allows one to create general models with less amount of effort for task-specific text classification. In this work, we show that bidirectional transformers can produce competitive results even without fine-tuning by leveraging auxiliary sentiment and emotion information \cite{DragutYSM10}.  Experimental results show the effectiveness of our model and its remarkable performance. The model has a significantly smaller size compared to the BERT-base model. \par

The main contributions of this work are as following:
\begin{itemize}
    \item Proposing a simple but efficient recurrent neural network that leverages sentence-wise sentiment or token-level emotion of input sequence with BERT representation for detecting the stance of a long perspective against its related question.
    \item Creating a novel dataset for stance detection with more than $6$K instances.
    \item Explaining the word/phrase contribution of input sequence using max-pooling engagement score for stance detection.
\end{itemize}

\section{ Related Works}
We group stance detection methods based on underlying data and  approaches as follows:
\begin{itemize}
    \item \textit{Tweets} are collected from SemEval 2016, Task 6, \cite{mohammad-etal-2016-semeval} and organized in two categories.  The first category,
    which represents a supervised setting, includes tweets that cover opinions about five     topics, 
    ``Atheism", ``Climate Change", ``Feminist Movement", ``Hillary Clinton", and ``Legalization of Abortion". The second category, which represents weakly supervised settings, includes tweets that cover one topic, but the training data is unlabeled. 
    \item \textit{Claims} are obtained from Wikipedia in \cite{bar-haim-etal-2017-stance}. Each claim is defined as a brief statement that is often part of a Wikipedia sentence. The claim dataset contains $55$ different topics. 
   \item  \textit{Debates} are gathered from various online debate resources, including \texttt{idebate}, \texttt{debatewise} and \texttt{procon} in the form of perspective, claim, and evidence for substantiated perspective discovery.  49 out of its 947 claims are from \texttt{procon} \cite{chen-etal-2019-seeing}. Claims and perspectives are short sentences and have been used for stance detection in \cite{Popat2019STANCYSC}.
 
\end{itemize}
Current approaches on stance detection use different types of linguistic features, including word/character n-grams, dependency parse trees, and lexicons \cite{sun-etal-2018-stance,sridhar-etal-2015-joint,hasan-ng-2013-stance,walker-etal-2012-stance}.
 There are also end-to-end  neural network approaches that learn topics and opinions independently while joining them with memory networks \cite{mohtarami-etal-2018-automatic},
bidirectional conditional LSTM \cite{augenstein-etal-2016-stance}, or neural attention  \cite{du2017stance}. There are also some neural network approaches that 
leverage lexical features \cite{riedel2017fnc,hanselowski-etal-2018-retrospective}. A consistency constraint is proposed to jointly model the topic and opinion using BERT architecture \cite{Popat2019STANCYSC}. It trains the whole massive network for label prediction.  None of these approaches incorporate bidirectional transformers with sentiment and emotion in a shallow neural network as we propose in this paper. Additionally, our focus is to find the stance of 100-200 words long discussions, which are commonly present in nonpartisan format.

\section{Dataset}
We collect data from \url{procon.org}, a non-profit organization that presents opinions on controversial issues in a nonpartisan format. Issues (questions) and their related responses are professionally researched from different online platforms by its experts. The dataset covers $419$ different detailed issues ranging from politics to sport and healthcare. 
The dataset instances are pairs of issues, in the form of \textit{questions},  and their corresponding \textit{perspectives} from proponents and opponents. Each  \textit{perspective} is either a pro or a con with 100-200 words that supports its claim with compelling arguments. 
Table \ref{tab:examples} provides some examples of the questions from the dataset. The dataset statistics are also presented in Table  \ref{tab:dataset}. We may use the words \textit{opinion} and \textit{perspective} interchangeably as both refer to the same concept in this work.

\begin{table}[t]
    \centering
    \begin{tabular}{|l|}
    \hline
    HEALTH and MEDICINE\\
    1- Should euthanasia or physician-assisted\\ suicide be    legal?\\
    2- Is vaping with e-cigarettes safe?\\
    \hline   
    EDUCATION\\
    
        1-Should parents or other adults be able to\\
        ban books from schools and libraries?\\
        2- Should public college be tuition-free?\\
        \hline
        POLITICS\\
        1- Should recreational marijuana be legal?\\
        2- Should more gun control laws be enacted?\\
        \hline
        SCIENCE and TECHNOLOGY\\
        1- Is cell phone radiation safe?\\
        2- Should net neutrality be restored?\\
        \hline
        ENTERTAINMENT and SPORTS\\
        1- Are social networking sites good for \\our society?\\
        2- Do violent video games contribute to\\ youth violence?\\
        \hline
    \end{tabular}
    \caption{Procon dataset questions}
    \label{tab:examples}
\end{table}

\begin{figure*}
\centering
  \includegraphics[width=0.6\textwidth]{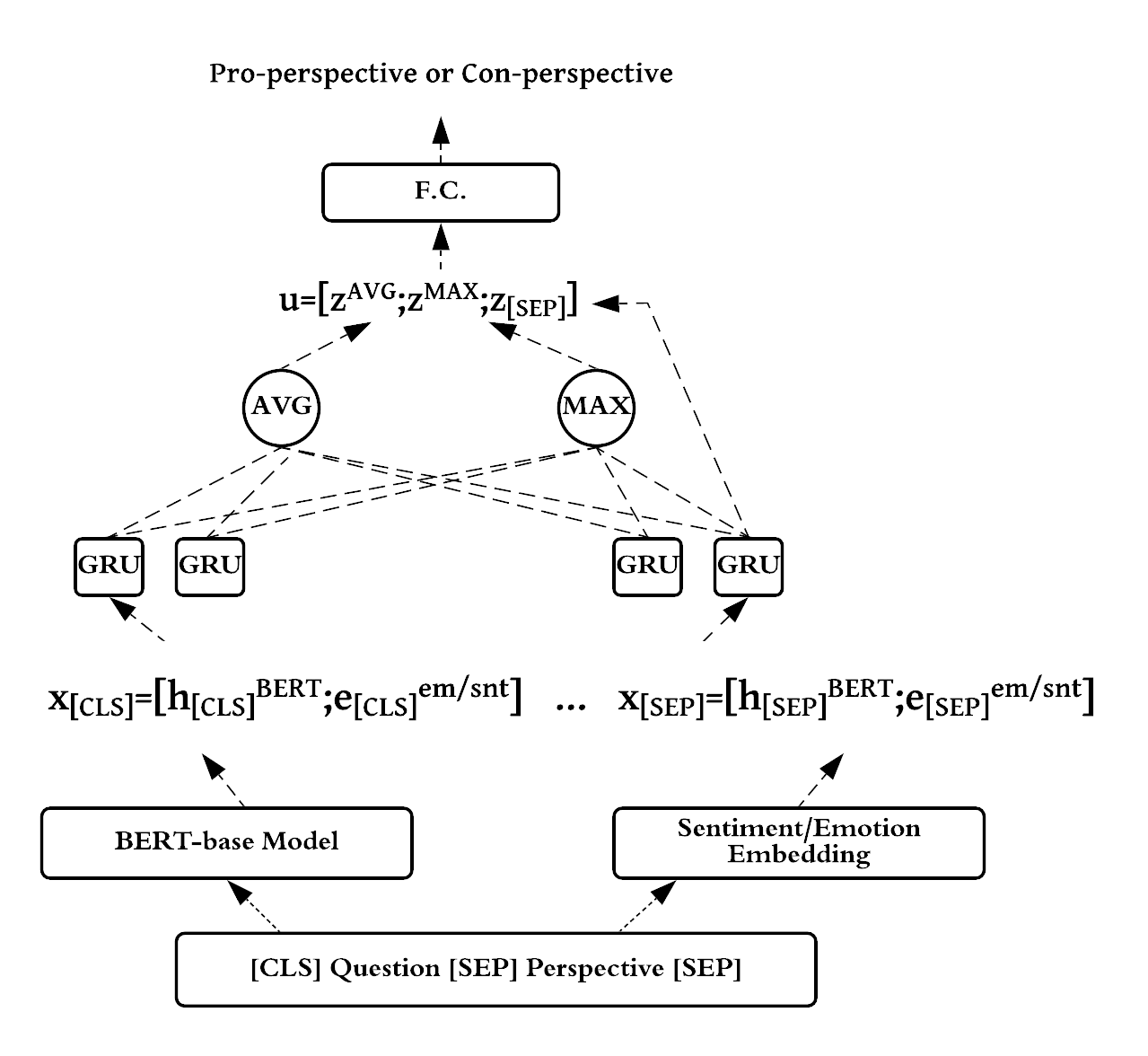}
    \caption{\label{fig:architecture} Stance detection architecture; snt: sentiment; em: emotion }
    
\end{figure*}

\begin{table*}[t]
  \centering
  
    \begin{tabular}{|l|l|l|l|l|l|}
    \hline
       Set   &  \#of Topics & \#of Words  & \#of Pro-perspectives  & \#of Con-perspectives  & Total \\
    \hline
    train & 417  & 127    &   2,140    & 2,125 &  4,265\\
    dev   & 265 & 125     &  326    &  284 & 610\\
    test  &  336  &123   &   613    & 606 & 1,219\\
    \hline
    \end{tabular}%
  \caption{\label{tab:dataset} Procon dataset statistics}
\end{table*}%

\section{Model}
\label{model}
Utilizing pre-trained models has been widely popular in machine translation and various text classification tasks. Prior efforts were hindered by the lack of labeled data \cite{ZhangHDV19}.
With the growth of successful pre-trained models, a model fine-tuned on a small portion of data can compete with models trained on 10$\times$ more training data without pre-training \cite{howard-ruder-2018-universal}. 
Recently, transformer models trained on both directions of language simultaneously,  such as BERT \cite{devlin-etal-2019-bert} and XLNet \cite{yang2019xlnet}, overcome previous unidirectional language models (e.g. ULMFiT \cite{howard-ruder-2018-universal}) or models trained on  two independent directions (ELMo) \cite{peters-etal-2018-deep} significantly.
So, we build our baselines based on BERT architecture in two different ways: single and pair of inputs. A question and its related opinion are concatenated for single inputs.  However, for input pairs, the question and the opinion are being separated with the BERT separator tag [SEP]. This approach has been used for question-answering applications \cite{devlin-etal-2019-bert}.

Opinion is connected with sentiment and emotion \cite{SchneiderD15}. Moreover, prior efforts show the successful employment of linguistic features, extracted with external tools, in neural networks for emotional cognition \cite{yang-etal-2017-satirical}. So, we leverage sentiment and emotion information separately with BERT representations obtained from the last BERT-base layer to form the input of a shallow recurrent neural network. In the following, we provide the details.

\begin{itemize}
    \item Employing sentiment:
We analyze how the sentiment of sentences in proponents' and opponents' opinions can affect stance detection. Accordingly, we use a rule-based sentiment tool,  VADER \cite{Hutto2014VADERAP},  for obtaining the sentiment of a sentence. VADER
translates its compound sentiment score,  ranging from $-1$ to $+1$, into negative sentiment labels for scores $\leqslant -0.05$, positive labels for scores $\geqslant +0.05$, and neutral for the scores between $-0.05$ and $+0.05$.


Here, we compute sentence-wise sentiment using VADER to let the model learn the flow of sentiment across the opinion. So, each token borrows the sentiment of its corresponding sentence. Then, an embedding layer converts the discrete labels into  $d$-dimensional vectors $(d=768)$  using a randomly initialized matrix $W^s_{3\times d}$; These representations are concatenated with BERT token embeddings to form the bidirectional Gated Recurrent Units ($\overrightarrow{\overleftarrow{\text{GRU}}}$)  input ($x_t$ for token $t$):
    
    \begin{equation*}
    \begin{aligned}
        x_t  &=  [h_{t}^{\text{BERT}};e_t^{\text{snt}}],\\
        z_{t}  &=  \overrightarrow{\overleftarrow{\text{GRU}}}(x_t),\\
        u   &=  [\text{avg-pool}(Z);\text{max-pool}(Z);z_T],\\
        y  &=   \text{softmax}(Wu+b)  
    \end{aligned}
    \end{equation*}
    
    For an input sequence with $T$ tokens, $h_{t}^{\text{BERT}}$ is the hidden state of the last  BERT-base layer corresponding to the input token at time  $t$, $e_t^{\text{snt}}$ is sentiment embedding of the token, [;] denotes concatenation operator, $Z=[z_i]_{i=1}^T$, and $W,b$ are parameters of a fully connected layer.
    
    Recall that our task is to identify the stance of long opinions; So, important information towards the final stance might be anywhere in the opinion. Because of that, we collect such information from the recurrent hidden states of all input tokens using max and average-pooling. Max-pooling returns a vector with maximum weights across all hidden states of input tokens for each dimension. In this way, the input tokens with higher weights will be engaged for stance prediction. Aside from that, the last hidden state of the recurrent network ($z_T$) is concatenated with the pooled information ($u$). Finally, a dense layer transforms vector $u$ into the class dimension. Figure \ref{fig:architecture} shows the model architecture.

    We refer to this model as VADER-Sent-GRU and report the experimental results in Section \ref{experiment}.
    
    \item Employing emotion: We take a similar approach to engage emotion information for stance detection using the NRC emotion lexicon \cite{Mohammad13}.  The Lexicon is collected by crowdsourcing and consists of English words with their eight basic emotions including anger, fear, anticipation, trust, surprise, sadness, joy, and disgust. So, the
    GRU input is a concatenation of BERT representation with emotion embedding (gained from a $9\times d$ matrix with random initialization; one dimension is added for neutral emotion). Here, we use unidirectional $\overrightarrow{\text{GRU}}$ as it shows more stable results in our pilot experiments.
    \end{itemize}

\section{Experiments}
In this section, we describe the corresponding  baselines  followed by the training setup.
\subsection{Baselines}
We use the following baselines utilized in opinion mining including sentiment analysis and stance detection: 

\begin{itemize}
    \item BERT \cite{devlin-etal-2019-bert} followed by a nonlinear transformation on a dense layer is used for downstream stance detection. Here, the whole network is fine-tuned and  all  $12$ BERT-base layers' weights will be updated in backpropagation. The information is pooled from the final hidden state of the classification token ($h_{\text{[cls]}}^{\text{BERT}}$)  after passing a fully connected layer with non-linear activation ($\tanh$). Then, a classifier layer shrinks the activations to a binary dimension.
   \begin{equation*}
    \begin{aligned}
        x  & =  \tanh(W^{p}h_{\text{[cls]}}^{\text{BERT}}+b^{p}), \\
   y & =  W^{c}x+b^{c}
       \end{aligned}
    \end{equation*}

where $W^c,W^p,b^p,\text{ and }b^c$ are the layers' parameters.

\item BERT$_{\text{CONS}}$ is a BERT base model that considers two different inputs using a perspective and its respective claim \cite{Popat2019STANCYSC}. The first input is similar to BERT sentence model's, \textit{[CLS] claim [SEP] perspective [SEP]}, and the second one is the sequence of \textit{[CLS] claim [SEP]}. Each input will be given to the BERT model separately. The goal is to incorporate the consistency between the representation of the perspective and claim using cosine distance of the two inputs.  Accordingly, the following loss (loss$_c$) is added to the regular cross-entropy loss of the BERT model:

\begin{equation*}
    \begin{aligned}
    loss_c=
    \begin{cases}
         1- cos(X^{[C]}, X^{[C;P]}), \text{y=pro}  \\
         max(0,cos(X^{[C]}, X^{[C;P]}), \text{y=con}  \\
    \end{cases} 
    \end{aligned}
\end{equation*}

where $X^{[C]}$ and $X^{[C;P]}$ are the final hidden state representations corresponding to the [CLS] token of the BERT model for the specified input. In our experiments, we replace the underlying question of a perspective with the claim in the two input sequences.

    \item XML-CNN model consists of three convolution layers with kernel size$=(2,4,8)$. With a  dynamic max-pooling layer, crucial information is extracted across the document. XML-CNN was able to beat most of its deep neural network baselines in six benchmark datasets \cite{liu2017deep}. We use, BERT, Word2vec, and FastText \cite{mikolov2018advances} embeddings for input tokens.
    \item AWD-LSTM  is a weight-dropped LSTM that deploys DropConnect on hidden-to-hidden weights as a form of recurrent regularization \cite{merity2017regularizing}.  Word2vec Embedding is used for its input. 

\end{itemize}

 We define the corresponding hidden states of the last BERT layer as BERT embedding/representation of input sequence for both single and pair of inputs mode.

\subsection{Training}
We develop our code based on the Hedwig\footnote{https://github.com/castorini/hedwig}  implementation and train the models on $30$ epochs with batch size=$8$. We apply early stopping technique to avoid overfitting during training.  Training is stopped after $5$ consequent epochs of no improvement of the highest F1 score.
 We inspect the test set on the model with the best F1 score of development set and keep the settings for BERT the same as the \textit{BERT-base-uncased} model. Adam optimizer with the learning rate of $2e-5$ (for BERT)  and $2e-4$ (for other models) is used. We see a dramatic drop in BERT performance with some other learning rates. Scikit-learn \cite{scikit-learn} library is employed for evaluation measures.

\begin{table}[t]
  \center
  \resizebox{0.48\textwidth}{!}{
    \begin{tabular}{|l|l|l|l|}
   \toprule
    
    Model &    P. & R. & F1   \\
    \hline
  \multicolumn{4}{|c|}{Pair of Input}\\

    \hline

 BERT &  \textbf{76.49}   &  75.37   & $75.92^{\ast \dagger}$\\
 \hline
 BERT$_\text{CONS}$ & 70.34 &  81.24 &  75.40\\
  \hline

XML-CNN(BERT) &  68.48  &  82.22 &  74.72 \\

   \hline   
  \textbf{VADER-Sent-$\overrightarrow{\overleftarrow{ \text{GRU}}}$} &  69.14  & \textbf{86.62}  & \textbf{76.90$^{\dagger}$}\\
  \hline
  NRC-Emotion-$\overrightarrow{\text{GRU}}$ &   73.79 &  79.45 &  \colorbox{lightgray}{$76.51^{\ast}$}  \\
  \hline
  
  \multicolumn{4}{|c|}{Unary Input}\\

      \hline
 BERT & \colorbox{lightgray}{73.89} &  76.18 &  75.02  \\
  \hline

AWD-LSTM(Word2Vec) & 65.93  & 73.25 &  69.40 \\

  \hline
  XML-CNN(Word2Vec) &58.30  & 83.03 &  68.51 \\
  \hline
  XML-CNN(FastText) & 66.85 &  77.32  & 71.71 \\
  \hline
  XML-CNN(BERT) & 70.30  & 79.93  & 74.81 \\
  \hline
 VADER-Sent-$\overrightarrow{\overleftarrow{\text{GRU}}}$ &  66.36    & 82.38   & 73.51\\
 \hline
 NRC-Emotion-$\overrightarrow{\text{GRU}}$ & 68.46  &  \colorbox{lightgray}{83.20}  &  75.11  \\
  \bottomrule
    \end{tabular}%
    }
 
  \caption{\label{tab:res} Evaluation results; P.:Precision, R.:Recall,  $^{\ast}$: $\text{\emph{p}}$-value  $\leq 0.001 $; $^{\dagger}$: $\text{\emph{p}}$-value  $\leq 0.0001 $ }
\end{table}%

\begin{table*}[!htbp]
    \centering
    \begin{tabular}{|l|l|l|l|}
    \toprule
       Model & P. & R. & F1  \\
        \hline
        &&&\\
        VADER-Sent-$\overrightarrow{\overleftarrow{\text{GRU}}}$ & 69.14  & 86.62  & 76.90\\
        $\overrightarrow{\overleftarrow{\text{GRU}}}$ & 72.18 &  78.30 &  75.12 $(\boldsymbol{1.78\downarrow})$ \\
        \hline
        &&&\\
        NRC-Emotion-$\overrightarrow{\text{GRU}}$ &   73.79 &  79.45 &  76.51\\
        
        $\overrightarrow{\text{GRU}}$ & 69.14 &  83.69  & 75.72 $(\boldsymbol{0.79\downarrow})$\\
        
       \bottomrule
    \end{tabular}
    \caption{Effect of sentiment and emotion in our models with pair of input}
    \label{tab:sentiment}
\end{table*}

\begin{table*}[t]
  \center
  
    \begin{tabular}{|ll|}
    \toprule
    \multicolumn{2}{|c|}{
\includegraphics[width=\textwidth]{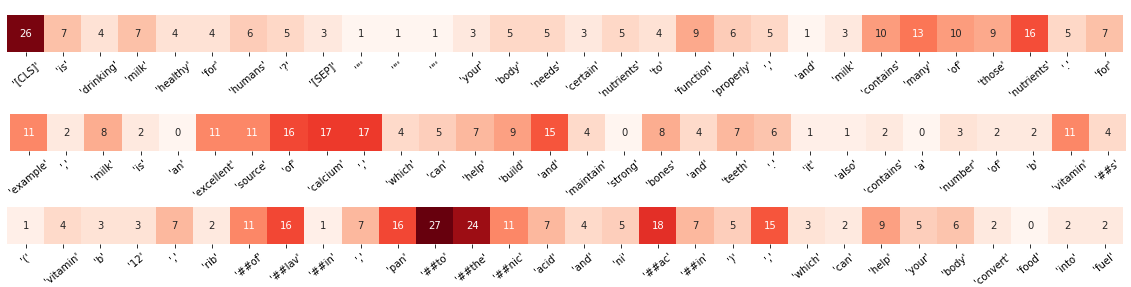}
}
 \\
 \hline
 Question : &  Is drinking milk healthy for humans?\\
 Top words: &  nutrients, calcium,  niacin, riboflavin, and pantothenic \\
 \bottomrule
    \end{tabular}
       \caption{A pro-perspective }
    \label{table:pro}
    \end{table*}

\begin{table*}[!htbp]
  \center
  
    \begin{tabular}{|ll|}
    \toprule
    \multicolumn{2}{|c|}{
\includegraphics[width=\textwidth]{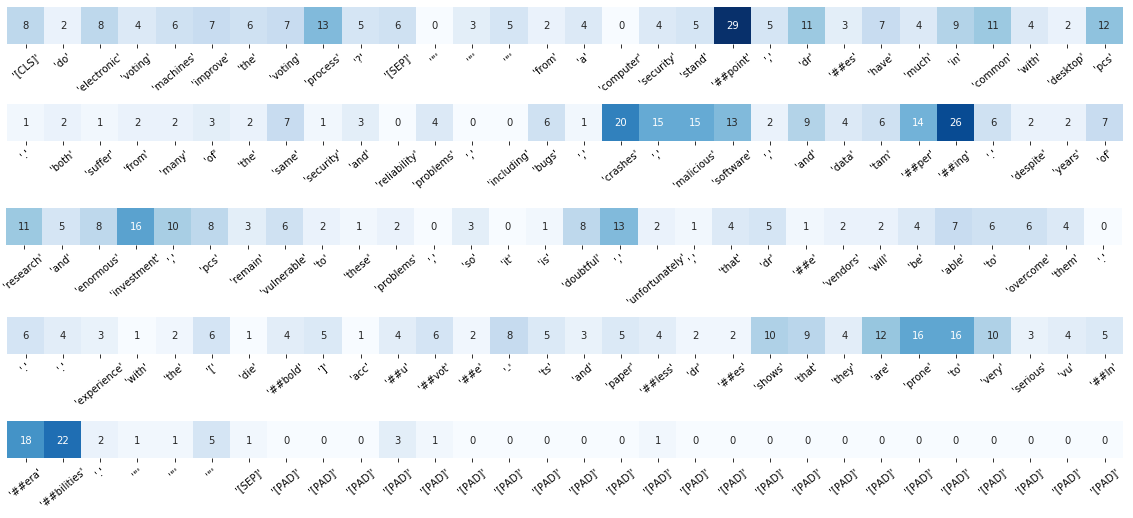}
}
 \\
 \hline
 Question : & Do electronic voting machines improve the voting process?  \\
 Top words: &  vulnerabilities, investment, standpoint, crashes, malicious software, and tampering  \\
 \bottomrule
    \end{tabular}
       \caption{A con-perspective }
    \label{table:con}
    \end{table*}

\section{Results and Discussion}
\label{experiment}
Experimental results are provided in Table \ref{tab:res}. It was expected that fine-tuning BERT with a pair of input achieves a competitive performance among other baselines; but it shows that even with a shallow concatenation of the question and perspective (unary input), BERT can achieve consistent results. Moreover, models that take BERT representation in feature selection mode (without fine-tuning), e.g. XML-CNN(BERT), show better stance detection performance than other token embeddings.

 We apply McNemar's test to measure whether the disagreement between the predictions of the two models is statistically significant.
 
 Among the models with pairs of input, VADER-Sent-GRU gains the highest recall and F1 score. It indicates that the external knowledge gained from a massive corpus, fine-tuned on $20\times$ fewer parameters and enriched with sentiment information can compete with the original architecture ($75.92$ vs $76.90$, $\text{\emph{p}}< 0.0001$ ). As the model is significantly smaller, it trains faster and needs fewer resources for training.
NRC-Emotion-GRU, highlighted in gray, achieves the second-highest F1 score among the models; It reveals that adding emotion information improves stance detection ($75.92$ vs $76.51$, $\text{\emph{p}} < 0.001$). However, employing sentiment information is more helpful than emotion in detecting the stance of opinions with compelling arguments ($76.51$ vs $76.90$, $\text{\emph{p}} < 0.0001$).

Unlike the superiority of BERT$_\text{CONS}$ over BERT reported in \cite{Popat2019STANCYSC}, we do not see a similar performance here. BERT$_\text{CONS}$ uses cosine similarity between the BERT representations of [claim] and  [perspective; claim] in the loss function such that their representations become similar when perspective supports the claim and dissimilar when it opposes the claim. This method works for claims and perspectives of the Perspectrum dataset where the two input components are short sentences with  $5-10$ words long. However, in our dataset, we have a \textit{question} and its perspective that spans multiple sentences. So, forcing the model to make the BERT representations of [question] and  [perspective; question] similar or dissimilar, according to the stance, harms the model training. Because the input components have different characteristics utilizing this method results in lower performance than the base model (BERT).

Next, we present some experiments to better understand the model's units.
\begin{figure*}
\centering
  \includegraphics[width=0.6\textwidth]{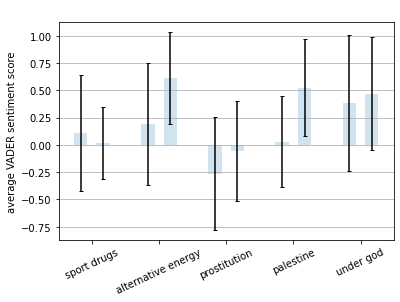}
    \caption{\label{fig:sentiment} Average VADER sentiment scores across five different issues. In each issue the first bar belongs to proponents and the second bar belongs to the opponents}
    
\end{figure*}

\subsection{Effect of Sentiment and Emotion}
As stated in Section \ref{model}, our recurrent models (VADER-Sent-GRU and NRC-Emotion-GRU) employ sentiment and emotion information of tokens respectively.  To see the effect of learning the flow of sentiment and emotion across an opinion, we lift their embeddings from the input of the models. So, $\overrightarrow{\text{GRU}}$  and $\overrightarrow{\overleftarrow{\text{GRU}}}$  are  unidirectional and bidirectional  Gated Recurrent Units network respectively,  followed by pooling and classification layers:

\begin{equation*}
    \begin{aligned}
                 x_t  &= h_{t}^{\text{BERT}},\\
        z_{t}  &=  \text{GRU}(x_t),\\
        u   &=  [\text{avg-pool}(Z);\text{max-pool}(Z);z_T],\\
       y  &=   \text{softmax}(Wu+b)  
    \end{aligned}
    \end{equation*}

    Similarly, for an input sequence with $T$ tokens, $h_{t}^{\text{BERT}}$ is the hidden state of the last BERT layer corresponding to the input token at time  $t$, $Z=[z_i]_{i=1}^T$, and $W,b$ are parameters of a fully connected layer.

According to the results in Table \ref{tab:sentiment}, both precision and F1 score reduce for the model without emotion ($\overrightarrow{\text{GRU}}$); however, we see a reduction in recall and F1 in the model after lifting sentiment ($\overrightarrow{\overleftarrow{\text{GRU}}}$) indicating that integrating sentence-wise sentiment and token-level emotion impact stance detection. We also provide the average sentiment score of the perspectives regarding five different questions in Figure \ref{fig:sentiment}. The figure shows the difference between the sentiment of the two stance classes in each issue resulting in a better stance classification. In the next part, we analyze the effect of pooling.


\subsection{Pooling Explanation}
 In \cite{Popat2019STANCYSC}, authors find the most important phrases of input by removing phrases from the sequence and finding the ones with maximum effect on misclassification. In our model, we find the crucial information engaged in identifying the stance of a perspective using the max-pooling operation applied to the output sequence of recurrent neural networks (see Section \ref{model}). We hypothesize that the more a token is engaged in max-pooling, the more critical the token is for final stance prediction.

 Tables \ref{table:pro} and \ref{table:con} show the heatmap plots of two test instances. The number in each square is the engagement score, the frequency of the presence of a token in max-pooling operation. Darker colors show a higher frequency and indicate how the model identifies the stance across the perspective towards a question.
 The underlying question in Table \ref{table:pro} asks `Is drinking milk healthy for humans?' According to its figure, we find sub-tokens of \textit{nutrients}, \textit{calcium},  \textit{niacin}, \textit{riboflavin}, and \textit{pantothenic} with high scores. All of these  words are positively aligned with the final (pro) stance; Specifically, the last three words are a type of Vitamin B.  In another example in Table \ref{table:con}, the question is `Do electronic voting machines improve the voting process?'  Its corresponding heatmap displays sub-tokens of  \textit{vulnerabilities}, \textit{investment}, \textit{standpoint}, \textit{crashes}, \textit{malicious software}, and \textit{tampering} with high scores; all of which are almost consistent with the perspective's (con) stance.  
 
  Similarly, we find the most important words/phrases, regarding their engagement score, for a few other examples of the test set that are correctly classified. The sub-tokens of these phrases have the highest frequency in max-pooling operation. We add (pro) or (con)  at the end of each phrase list to indicate the stance of their respective perspective.

 \begin{itemize}
     \item  Should students have to wear school uniforms?    uniforms restrict students' freedom of expression (con)
     \item        Are social networking sites good for our society?  lead to stress and offline relationship (con)
     \item        Should recreational marijuana be legal? legalization,  odious occasion (con)
     \item What are the pros and cons of milk's effect on cancer?   dairy consumption is linked with rising death rates from prostate cancer (con)
     \item Is human activity responsible for climate change?  significant, because,  (likely greater than 95 percent probability) (pro)
     \item Is obesity a disease?  no question that obesity is a disease, blood sugar is not functioning properly, dysregulation, diabetes (pro)
     \item Is the death penalty immoral?  anymore, failed policy (pro)

 \end{itemize}

The above list shows that the stance-related phrases have been well identified by the model in the pooling step.

\section{Conclusion}
We propose a model that leverages BERT representation with sentiment or emotion information for stance detection. We create a new dataset for the perspectives that are as long as a paragraph covering a wide variety of contemporary topics. The experiments on our benchmark dataset highlight the effect of emotion and sentiment in stance prediction. The model can improve BERT base performance with significantly fewer parameters. We also explain the contribution of essential phrases of perspectives in detecting their stance using max-pooling operation.

 \section*{Acknowledgments}
 This  work  is  supported  in  part  by  the U.S. NSF grants 1838147 and 1838145.  We  also  thank  anonymous reviewers for their helpful feedback.
 
 \bibliography{acl2020} 
\bibliographystyle{acl_natbib}
\end{document}